\documentclass{article}
\usepackage{spconf,amsmath,graphicx,diagbox}
\usepackage{color}

\makeatletter
\renewcommand{\maketag@@@}[1]{\hbox{\m@th\normalsize\normalfont#1}}%
\makeatother

\title{GeoGCN: Geometric Dual-domain Graph Convolution Network for Point Cloud Denoising}
\name{Zhaowei Chen$^1$, Peng Li$^1$, Zeyong Wei$^1$, Honghua Chen$^1$, Haoran Xie$^2$, Mingqiang Wei$^1$, Fu Lee Wang$^3$}
\address{$^1$Nanjing University of Aeronautics and Astronautics\\
$^2$Lingnan University\  \ $^3$Hong Kong Metropolitan University}

\begin{document}
%
\maketitle
\begin{abstract}

We propose \textbf{GeoGCN}, a novel geometric dual-domain graph convolution network for point cloud denoising (PCD). Beyond the traditional wisdom of PCD, to fully exploit the geometric information of point clouds, we define two kinds of surface normals, one is called Real Normal (RN), and the other is Virtual Normal (VN). RN preserves the local details of noisy point clouds while VN avoids the global shape shrinkage during denoising.
GeoGCN is a new PCD paradigm that, 1) first regresses point positions by spatial-based GCN with the help of VNs, 2) then estimates initial RNs by performing Principal Component Analysis on the regressed points, and 3) finally regresses fine RNs by normal-based GCN.
Unlike existing PCD methods, GeoGCN not only exploits two kinds of geometry expertise (i.e., RN and VN) but also benefits from training data. Experiments validate that GeoGCN outperforms SOTAs in terms of both noise-robustness and local-and-global feature preservation.

\end{abstract}
\section{Introduction}
\label{sec:intro}

Point clouds have seen growing popularity with wide applications of autonomous driving, and Metaverse in recent years \cite{li2020deep}. They are often acquired by advanced 3D sensors, where noise inevitably creeps in, due to measurement and reconstruction errors. Noise reduces the
accuracy of captured surfaces and should be reduced before
many geometric tasks. 


Given a measurement $\textbf{P}^{\star}=(\textbf{P}+\varepsilon)$ with noise $\varepsilon$, point cloud denoising (PCD) aims to recover a clean point cloud $\textbf{P}$ to represent the sampled underlying surface.
Current PCD methods are roughly divided into conventional methods and learning-based methods. The former may exploit certain surface assumptions, like sparsity or non-local similarity, to remove noise while attempting to preserve the surface's geometry \cite{lipman2007parameterization,guennebaud2007algebraic,dinesh2020point,chen2019multi}. However, they are time-consuming and require careful trial-and-error parameter tuning. Recent years have witnessed extensive efforts in learning-based methods \cite{roveri2018pointpronets,yu2018pu,yu2018ec}. They usually take a small patch reflecting the local geometry as input to encode the geometric information and predict a displacement for each point. Few learning-based PCD methods can benefit from the global geometry to enhance the performance of denoising. 

We propose a novel and effective PCD method that exploits geometry expertise and benefit from training data. Our method is inspired by the following two observations. 

(1) The local geometry helps smooth the local noisy points while retaining sharp features; the global geometry helps constrain the overall shape of noisy point clouds. 
Thus, the two types of geometry are potentially combined to complement each other to promote the PCD performance. Based on the aforementioned analysis, we define two types of normals as surface geometry signals: Real Normal (RN) and Virtual Normal (VN). RNs are locally estimated and utilized to retain the local details, and VNs are globally estimated and utilized to constrain the global shape. 

(2) A point cloud with noise removed can more accurately estimate its surface normals than its noisy version. Meanwhile, more accurate surface normals can better remove noise than their raw surface normals.
Thus, we design a cascaded two-stream network that first regresses point coordinates and then point normals, in which RNs and VNs are integrated. The regressed point coordinates can be used to estimate the initial RNs by PCA. 
The regressed RNs can be used to fit the new point coordinates to yield the final PCD result.

Our main contributions are as follows:\\
(1) We define two types of surface normals: RN is exploited to update the vertex coordinates to optimize the local details of noisy point clouds, and VN is used as a loss function to constrain the global shape of noisy point clouds.\\
(2) We propose a novel point cloud denoising paradigm based on a dual-domain graph convolutions, which learns the defined RNs and VNs from the spatial and normal domains and synthesizes RNs and VNs to improve the PCD performance.

\section{Method}
\label{sec:method}

\begin{figure*}
  \centering
  \includegraphics[width=16cm]{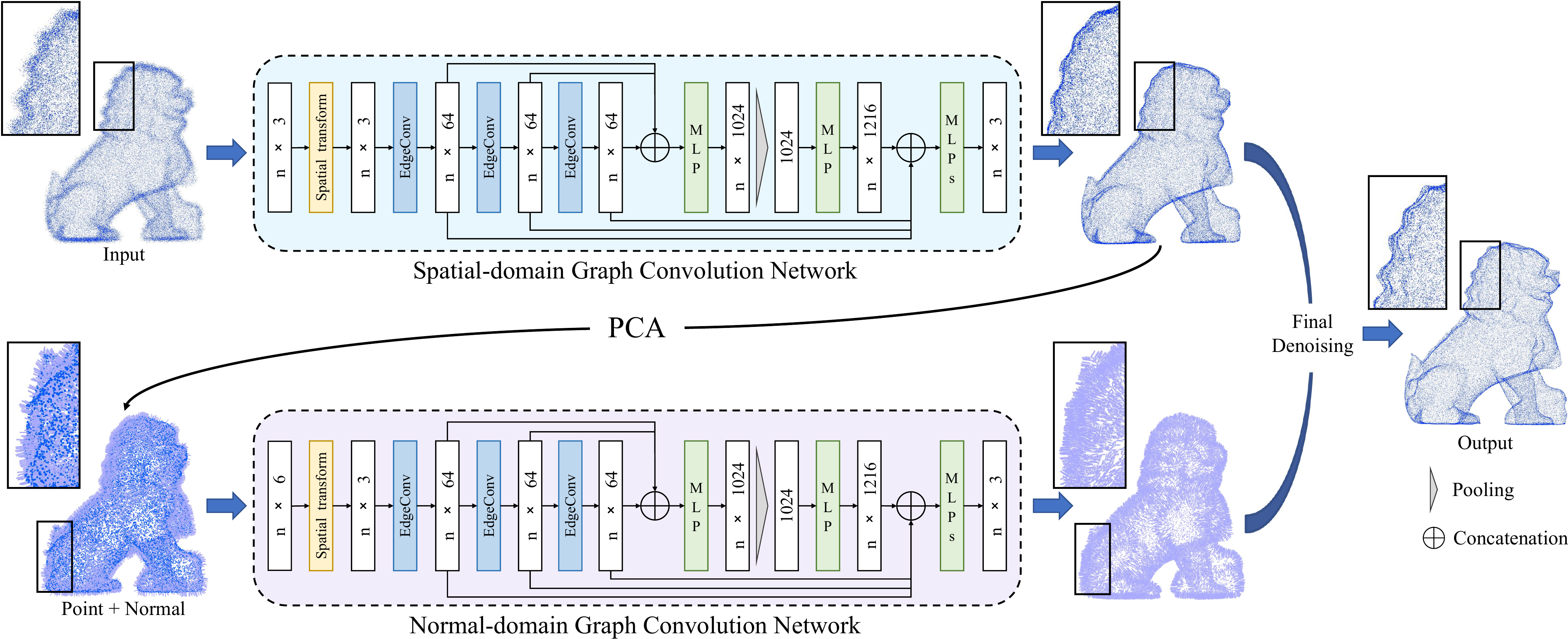}
  \caption{\textbf{The pipeline of GeoGCN}: The input of GeoGCN are patches of noisy points. The S-GCN regresses the initial denoised patches. Then, PCA is performed on the initial denoised patches to achieve initial RNs, which are concatenated with the initial denoised points to feed into N-GCN. The N-GCN is used to regress the fine RNs. The final denoising module generates the final denoised point clouds.}
  \label{pipeline}
\end{figure*}

\subsection{Overview}
\label{ssec:subhead}

Current network architectures seldom leverage geometry expertise. 
In our PCD method, we exploit two kinds of geometry expertise.
(1) We formulate RN and VN representing the local and global geometry respectively and naturally integrate them into Dynamic Graph Convolutional Neural Network (DGCNN) \cite{wang2019dynamic} to enhance the denoising results. 
(2) We establish both the ``point-to-normal" and ``normal-to-point" relationships. The two relationships make our method robust to noise and faithful to the underlying noise-free surface.

 GeoGCN consists of two parts (see Fig. \ref{pipeline}): Spatial-domain Graph Convolution Network (S-GCN) and Normal-domain Graph Convolution Network (N-GCN). First, S-GCN takes a patch of the noisy point cloud as input to regress the initial denoised patch. The defined VNs are utilized as the loss function to constrain the shape of the point cloud. Then, we perform principal component analysis (PCA) \cite{abdi2010principal} on the initial denoised patches to achieve initial RNs. Finally, we concatenate the initial denoised points and corresponding RNs and feed them to N-GCN to generate the fine RNs. The fine RNs are utilized to update the vertex coordinates of the initial denoised points to optimize the local details.





\subsection{S-GCN}

S-GCN is a sub-network in the spatial domain. We input a patch of the noisy point cloud into S-GCN to regress the initial denoised patch. Through two loss functions, we constrain the training process in both the global and local geometry.

\begin{figure}[htbp]
  \centering
  \includegraphics[width=6cm]{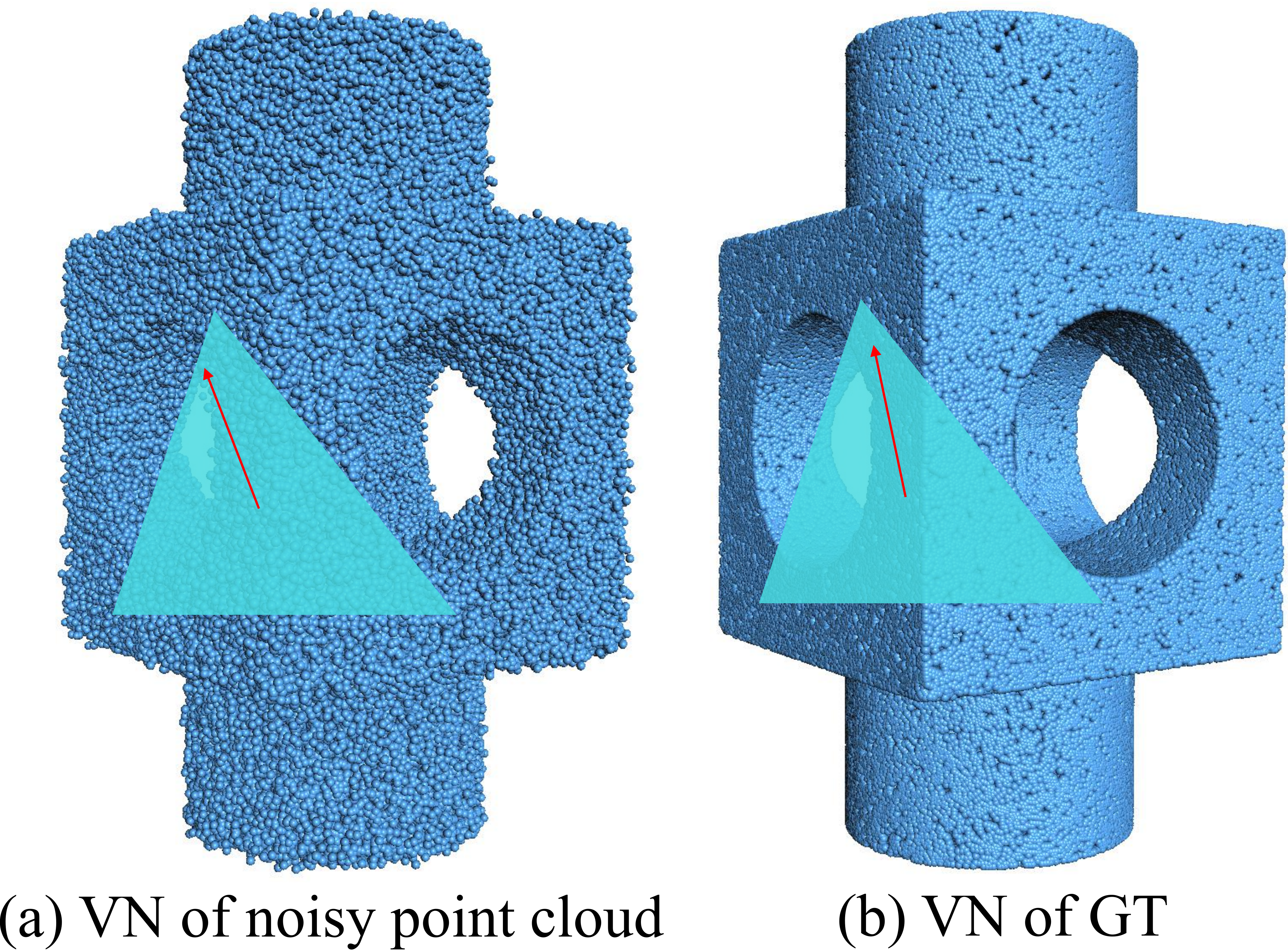}
  \caption{The schematic diagram of VN.}
  \label{VNs}
\end{figure}

\textbf{Virtual Normal.} We formulate Virtual Normal (VN) as the global geometry of a given point cloud. To this end, we construct a series of triangles by randomly picking up three points in the point cloud, and consider the normals of these triangles as the virtual normals (see Fig. \ref{VNs}).  
These VNs enable to preserve the global shape during denoising. 

Specifically, we randomly select N groups of three-points in the initial denoised point cloud outputted by S-GCN. 
We take two restrictions to ensure that VNs can well reflect the global geometry information of the point cloud. First, three points in each group should not form a line. We control the three angles of each triangle to avoid thin triangles. Thus, all three angles of any triangle range within $[45^{\circ}, 90^{\circ}]$. Second, the edge length of each triangle should be longer than a pre-fixed threshold, since a triangle with short edge lengths is sensitive to noise. 

\textbf{Loss.} After selecting $N$ groups of three-points satisfying the restrictions, we mark their indexes and find the corresponding points in the clean point cloud. Virtual Normal Loss is formulated to constrain the global shape of a point cloud as 
\begin{small}
\begin{equation}
Loss_{\mathrm{VN}}=\frac{1}{N}\sum_{i=0}^{N}{\left.|n_i^{pred}-n_i^{gt}\right.|} 
\end{equation}
\end{small}
where  $n_i^{pred}$ is VNs from the initial denoised points, and $n_i^{gt}$ is VNs from the clean point cloud. 

We also formulate Earth Mover Distance (EMD) \cite{hou2016squared} to regress coordinates in S-GCN as
\begin{small}
\begin{equation}
Loss_{\mathrm{EMD}}=EMD\left(P,Q\right)=\frac{\sum_{i=1}^{m}\sum_{j=1}^{n}{f_{ij}d_{ij}}}{\sum_{i=1}^{m}\sum_{j=1}^{n}f_{ij}}
\end{equation}
\end{small}
where $P$ and $Q$ represent the initial denoised point cloud outputted by S-GCN and the ground truth, respectively. $d_{ij}$ represents the distance from the $i^{th}$ point of $P$ to the $j^{th}$  point of $Q$. $f_{ij}$ indicates whether to move the $i^{th}$  point of $P$ to the $j^{th}$  point of $Q$. $EMD$ represents the total cost of moving all points in $P$ to all points in $Q$.

\subsection{N-GCN}

N-GCN is a sub-network in the normal domain. We concatenate the initial denoised points and corresponding RNs calculated by PCA, then feed them into N-GCN to generate the fine RNs. 

\textbf{Real Normal.} Principal Component Analysis (PCA) is employed to analyze
the covariance in a local structure around a noisy point. Real Normal is the eigenvector corresponding to the smallest eigenvalue of the covariance. Such normals can well reflect the local geometry of a point. 

\textbf{Loss.} PCA normals are un-oriented, and obtaining a consistent normal orientation requires additional effort.
Our solution is to either add or subtract the predicted RNs from the ground truth, in which the smaller value is adopted. Such an operation ensures that the regressed RNs are closer to the straight line with the ground truth. 
Thus, we formulate the $L_2$ loss in N-GCN to regress RNs as
\begin{small}
\begin{equation}
Loss_{\mathrm{RN}}=\sum_{i=1}^{n}{min\left(\left(n_i^{gt}-n_i^{pred}\right)^2,\left(n_i^{gt}+n_i^{pred}\right)^2\right)}
\end{equation}
\end{small}
where $n_i^{pred}$ represents RNs output by N-GCN and $n_i^{gt}$ is GT.

\subsection{Total Loss}
The total loss is formulated as 
\begin{small}
\begin{equation}
Loss=\alpha Loss_{\mathrm{EMD}} + \left(1-\alpha\right)Loss_{\mathrm{VN}} + \beta Loss_{\mathrm{RN}}
\end{equation}
\end{small}
where $\alpha$ and $\beta$ are the weights. We empirically set $\alpha$=0.9 and $\beta=0.1$ that behave well during denoising.

\subsection{Final Denoising}
To obtain more accurate point cloud denoising results, we utilize the regressed RNs to fit the regressed points. The final denoising point is generated by \cite{zhou2020geometry} 
\begin{small}
\begin{equation}
p_i^\prime=p_i+\gamma\sum_{p_j\in N_i}\left(p_j-p_i\right)\left(W_\sigma\left(n_i,n_j\right)n_i^Tn_i+\lambda n_j^Tn_j\right)
\end{equation}
\end{small}
where $p_i^\prime$ and $p_i$ are the point after and before denoising, $N_i$ is the neighboring points of $p_{i}$.  $W_\sigma\left(n_i,n_j\right)=\exp{\left(-\frac{\left.|n_i-n_j\right.|^2}{\sigma^2}\right)}$ is a weight function, $\lambda$=0.5, $\gamma=\frac{1}{3\left|N_i\right|}$ is the step size, and the number of iterations is set to 10.

\section{Results and Discussions}
\label{sec:ed}


\subsection{Dataset}
\label{ssec:subhead}

The training dataset consists of 22 point cloud models with accurate coordinates and normals, including 11 CAD models and 11 non-CAD models. Each model contains 100k points randomly sampled from its original surface. We add different scales of Gaussian noise to each model, with noise scales of 0.25\%, 0.5\%, 1\%, and 1.5\%, respectively. Thus, the training dataset consists of 88 point cloud models with noise, and 22 point cloud models with accurate coordinates and normals.

The test dataset consists of both synthesized noisy models and real-scanned models. For synthesized data, we add Gaussian noise at a scale of 0.5\%. In addition, we test 5 real scanned point clouds.

\subsection{Training Details}
\label{ssec:subhead}

Our GeoGCN is implemented by PyTorch. We train it on a PC equipped with an Intel 12700F CPU and an NVIDIA RTX-1080Ti GPU. We utilize the SGD optimizer with a batch size of 64. 8000 patches are selected in each model to feed into GeoGCN, with 1000 points for each patch. We set the training epochs and initial learning rate to be 10 and 1e-3 respectively. The learning rate decreases from 1e-3 to 1e-6 when the epoch increases. It takes about 15 hours to train our GeoGCN.

\subsection{Quantitative Comparison}
\label{ssec:subsubhead}

We compare the proposed GeoGCN with point cloud denoising methods including the traditional ones, i.e., WLOP \cite{huang2009consolidation}, CLOP \cite{preiner2014continuous}, RIMLS \cite{oztireli2009feature}, GPF \cite{lu2017gpf}, and the learning-based ones, i.e., PCN \cite{rakotosaona2020pointcleannet}, Pointfilter \cite{zhang2020pointfilter}, and TD \cite{hermosilla2019total}. 

To comprehensively evaluate GeoGCN, we calculate the Chamfer distance (CD) \cite{butt1998optimum} and Mean Square Error (MSE) \cite{allen1971mean} over the test dataset. As illustrated in Table \ref{quantitative}, our method achieves the lowest average errors.

\begin{table}[htbp]
\centering
\footnotesize
\caption{The average errors of CD and MSE of all methods.}
\begin{tabular}{|p{2.5cm}<{\centering}|p{1.5cm}<{\centering} p{1.5cm}<{\centering}|} \hline
\diagbox{Methods}{Metrics} &CD ($10^{-5}$) &MSE ($10^{-3}$)\\ \hline
Noisy   &3.656  &5.868	\\ 
WLOP	&1.898	&4.361  \\ 
CLOP	&1.537	&4.138  \\ 
RIMLS	&1.242	&4.173  \\ 
GPF	    &2.375  &4.896  \\
PCN	    &1.027	&4.075	\\ 
PF      &0.997  &3.986  \\
TD	    &1.576	&4.443  \\ 
Ours	&\textbf{0.987}	&\textbf{3.954}  \\ \hline
\end{tabular}
\label{quantitative}
\end{table}

\begin{figure*}[htbp]
  \centering
  \includegraphics[width=16cm]{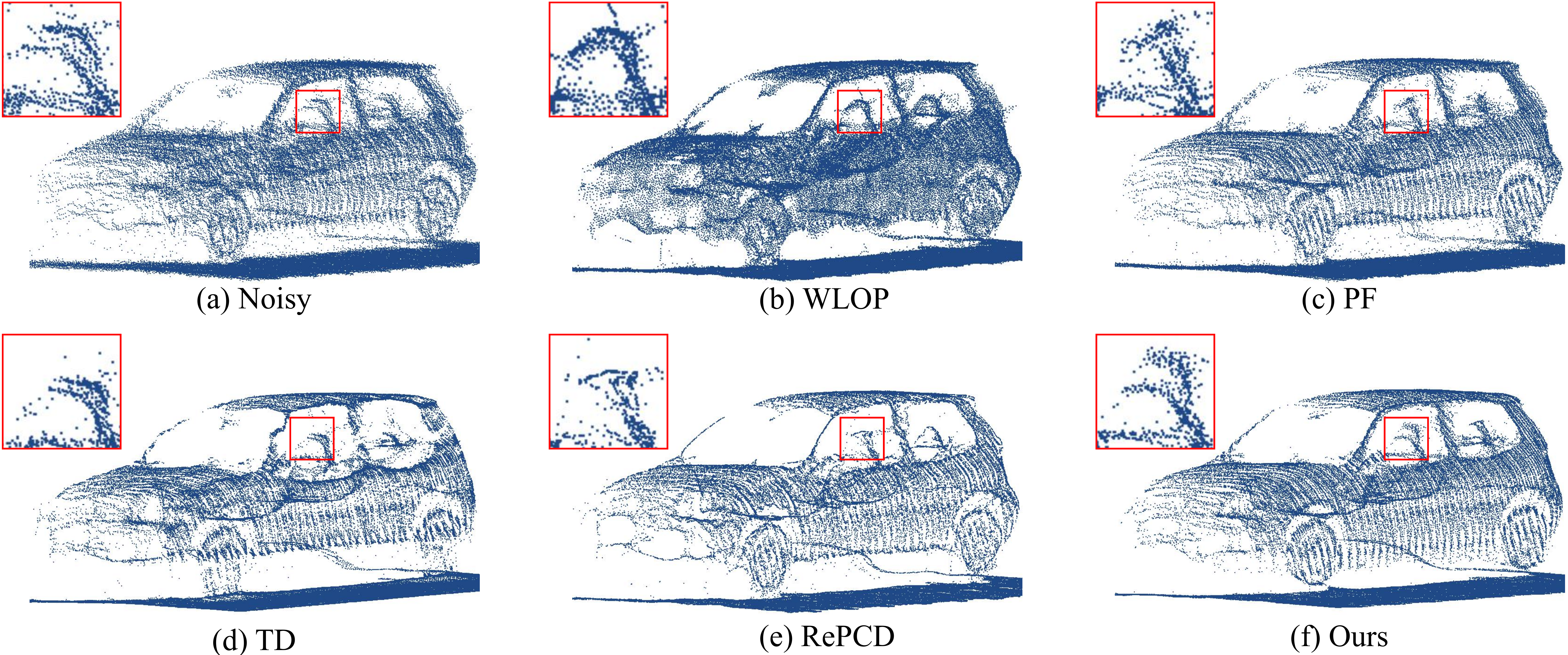}
  \caption{Denoising a real-scanned point cloud scene.}
  \label{visual}
\end{figure*}

\subsection{Visual Comparison}
\label{ssec:subhead}

We evaluate GeoGCN on raw scanned point clouds corrupted with raw noise. Since the ground-truth models of these raw scanned point sets are not available, we only demonstrate the visual comparisons with other methods but not with GT. As shown in Fig. \ref{visual}, we compare our method with WLOP \cite{huang2009consolidation}, Pointfilter \cite{zhang2020pointfilter}, TD \cite{hermosilla2019total}, and RePCD \cite{chen2022repcd}. As observed from these results, the headrest of our method is better than others. With the help of VNs and RNs, our method recovers the local details of the model, while optimizing the global shape.

\subsection{Ablation Study}
\label{ssec:subsubhead}

We perform ablation study to validate the effectiveness of VNs for global shape preservation and RNs for local shape preservation. GeoGCN is decomposed into three parts:\\
(1) S1: Learning the mapping from the noisy point clouds to the initial denoised points;\\ 
(2) S2: Adding Virtual Normal Loss;\\
(3) S3: Learning the mapping from the initial RNs to the fine RNs, and using fine RNs to update the initial denoised points.

\begin{figure}[htbp]
  \centering
  \includegraphics[width=8cm]{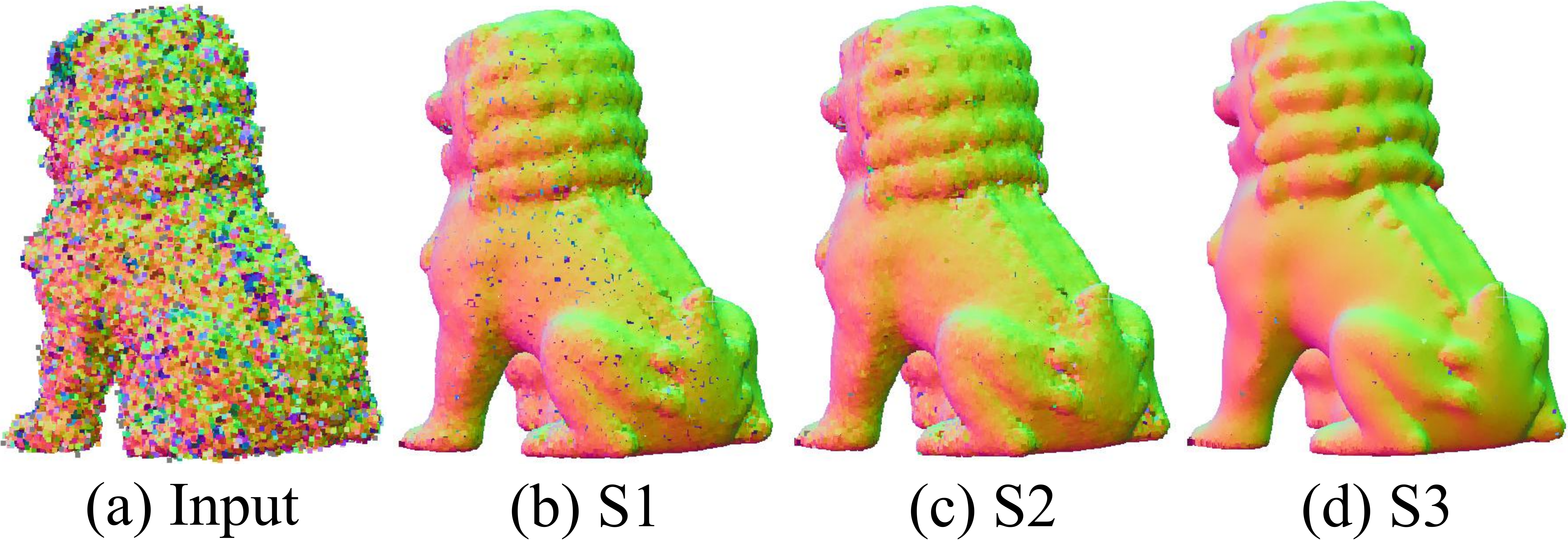}
  \caption{Denoising performance of GeoGCN.}
  \label{ablation image}
\end{figure}

As illustrated in Table \ref{ablation} and Fig. \ref{ablation image}, the average errors and visualizations of S1 are significantly better than that of the input, indicating that the initial denoised points from the first sub-network S-GCN are effective. Compared with S1, the average errors of S2 are smaller, indicating that VNs can improve the performance of the proposed model by preserving global geometric information. The average errors and visualizations of S3 are the best, indicating that utilizing the fine RNs regressed by the second sub-network N-GCN can effectively update the vertex coordinates of the initial denoised points and recover the local details.

\begin{table}[htbp]
\centering
\footnotesize
\caption{The average errors of MSE ($10^{-3}$).}
\begin{tabular}{p{1cm}<{\centering} p{3cm}<{\centering} p{3cm}<{\centering}} \hline
    &CAD     &non-CAD \\ \hline
Input &6.1111   &6.3123  \\
S1	&4.5687	 &4.8653  \\ 
S2	&4.2851  &4.5342  \\ 
S3	&\textbf{3.8976}	 &\textbf{4.0462}  \\ \hline
\end{tabular}
\label{ablation}
\end{table}

\section{Conclusion}
\label{sec:conclu}

Point cloud denoising is a fundamental yet not well-solved problem in 3D vision and graphics. Few efforts have been made to both benefit from the training data and absorb the geometric domain knowledge for point cloud denoising. 
We propose to formulate two kinds of normals as surface geometric signals, one is real normal to reflect the surface's local geometry and the other is virtual normal to reflect its global geometry. The two types of normals are flexibly integrated into current GCN architectures.
We consider that a noise-free point cloud can more accurately
estimate its surface normals than its noisy version; more accurate surface normals can better remove noise than their raw surface normals. Thus, we propose a cascaded
two-stream network that first regresses point coordinates in the spatial domain and
then point normals in the normal domain, in which RNs and VNs are integrated. Experiments show clear improvements of our method over its competitors in terms of noise-robustness and global-and-local shape preservation.

\vfill\pagebreak

\bibliographystyle{IEEEbib}
\bibliography{strings,refs}
\end{document}